%% file: Main.tex
\documentclass[runningheads]{llncs}

\usepackage{graphicx}
\usepackage{caption}
\usepackage{subcaption}
\usepackage[table,xcdraw]{xcolor}
\usepackage{algorithm}
\usepackage{algpseudocode}
\usepackage{multirow}
\usepackage{subfiles}
\usepackage{cite}
\usepackage{amsmath}

\usepackage[misc,geometry]{ifsym}
 
\usepackage[table]{xcolor} % loads also »colortbl«
\usepackage{sparklines}

\begin{document}
\title{RealCQA: Scientific Chart Question Answering as a Test-bed for First-Order Logic}
\titlerunning{RealCQA}

\author{
Saleem Ahmed(\Letter)\orcidID{0000-0001-8648-9625} ,\\
Bhavin Jawade, 
Shubham Pandey, 
Srirangaraj Setlur\orcidID{0000-0002-7118-9280}, 
Venu Govindaraju\orcidID{0000-0002-5318-7409}}
\authorrunning{
S. Ahmed et al.
}
\institute{University at Buffalo, USA \\
\email{\{sahmed9, bjawade,spandey8,setlur,govind\}@buffalo.edu}
}

\maketitle              % typeset the header of the contribution
\begin{abstract}
We present a comprehensive study of chart visual question-answering(QA) task, to address the challenges faced in comprehending and extracting data from chart visualizations within documents. Despite efforts to tackle this problem using synthetic charts, solutions are limited by the shortage of annotated real-world data. To fill this gap, we introduce a benchmark and dataset for chart visual QA on real-world charts, offering a systematic analysis of the task and a novel taxonomy for template-based chart question creation. Our contribution includes the introduction of a new answer type, `list’, with both ranked and unranked variations. Our study is conducted on a real-world chart dataset from scientific literature, showcasing higher visual complexity compared to other works. Our focus is on template-based QA and how it can serve as a standard for evaluating the first-order logic capabilities of models. The results of our experiments, conducted on a real-world out-of-distribution dataset, provide a robust evaluation of large-scale pre-trained models and advance the field of chart visual QA and formal logic verification for neural networks in general. Our code and dataset is publicly available \footnote{This is a pre-print version, accepted at ICDAR'23; https://github.com/cse-ai-lab/RealCQA}.

\keywords{Charts and Document Understanding and Reasoning}
\end{abstract}
\input{Sections/001_Introduction.tex}
\input{Sections/002_Background.tex}

\input{Sections/003_Methodology.tex}
\input{Sections/004_Results.tex}
\bibliographystyle{splncs04}
\bibliography{Main}
\end{document}

%% file: Sections/001_Introduction.tex
\section{Introduction}
The chart question-answering[QA] task has recently received attention from a wider community \cite{kafle2018dvqa}, \cite{methani2020plotqa}, \cite{chaudhry2020leaf}, \cite{singh2020stl}, \cite{masry2022chartqa}, \cite{levy2022classification}.
While generic multi-modal QA tasks have been studied widely, the chart-based QA task is still in its developmental phase, especially for real-world scientific document understanding applications.

Recent works have provided a structure for question type classification\cite{kafle2018dvqa} while iteratively adding complexity in chart types\cite{chaudhry2020leaf} and answer types \cite{methani2020plotqa}. However, no existing work fills the gap of QA on real-world charts\cite{icpr2022} with structured output prediction.

% The task involves a complex understanding of both text and visual components which have a semantic meaning to their layout(axis/legend/title etc). The visual complexity, the semantic nature of the data used to plot the charts, and language complexity in terms of the QA pairs - template-based, human-generated, etc need to be queried, understood, reasoned, and output either a closed set of two variables `Yes/No' , String Answers  with open vocab, Set based answers with ordered and un-ordered sets with a max of two items per set.

\begin{figure}[ht!]
         \centering
         \includegraphics[scale=0.6]{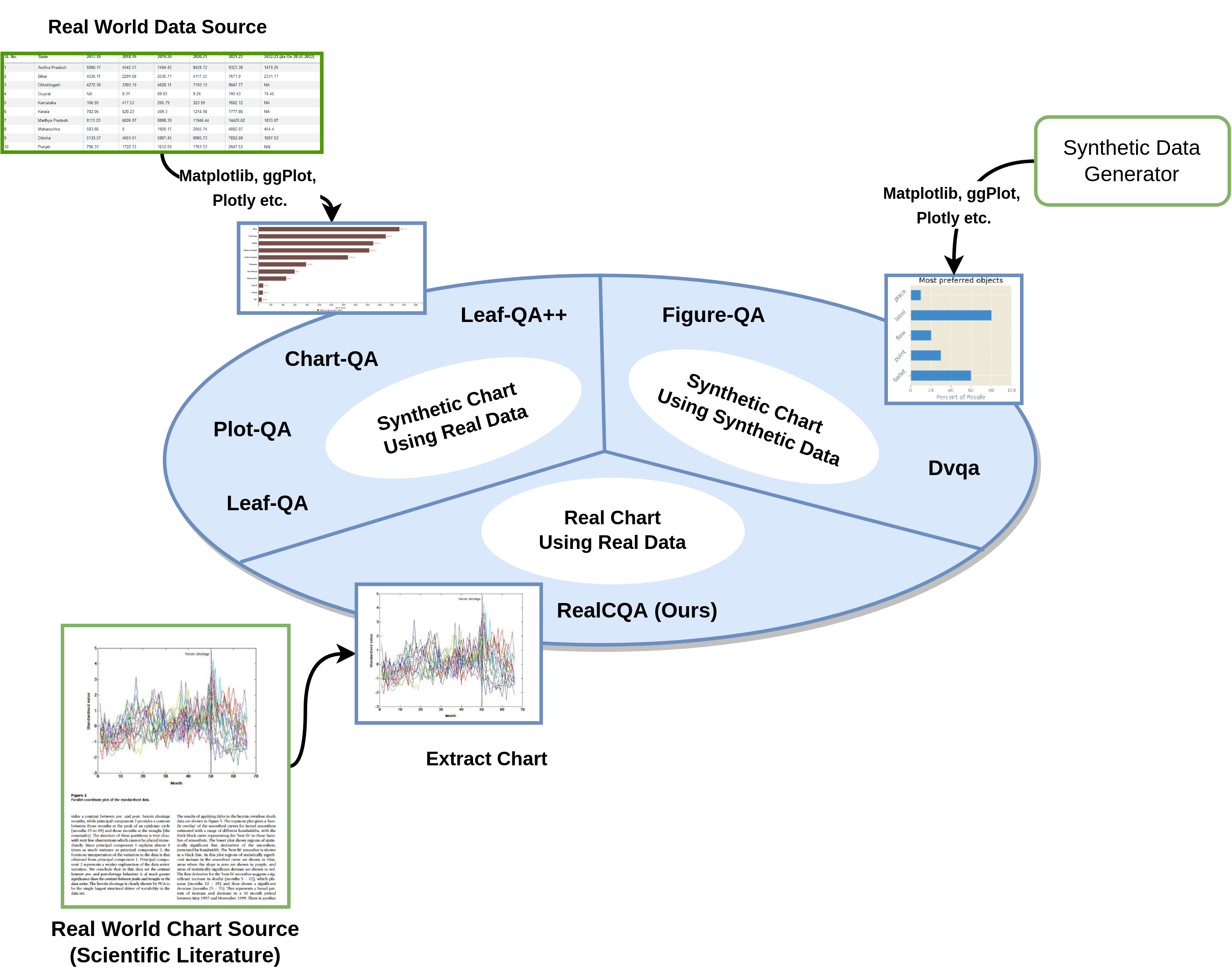}
         \caption{\scriptsize Existing datasets in chart visual QA either are fully synthetic charts generated from synthetic data (Right sector in the above ellipse) or synthetic charts generated from real data (left sector). None of these datasets handle the complexity of the distribution of real-world charts found in scientific literature. We introduce the first chart QA dataset (RealCQA) in the third category (lower sector in the above figure) which consists of real-world charts extracted from scientific papers along with various categories of QA pairs. [Best viewed digitally in color].}
         \vspace{-0.8em}
         \label{fig:existDatase}
\end{figure}

Two main approaches for synthetic Chart-QA are: (i) considering the whole input image as a matrix of pixels to generate output in the form of text, answer types, etc\cite{chaudhry2020leaf}, \cite{singh2020stl}, \cite{levy2022classification} or (ii) first extracting tabular data by identifying, classifying chart structural components and, then treating the task as a table-QA task\cite{masry2022chartqa}. 

These include either numeric answers (regression task) or single-string answers from the charts vocabulary (classification task). We further propose a structured and unstructured list answer type task, where answers can contain delimiter-spaced strings, where the order of strings might/or not matter. We also include new chart types for the scatter and box plots with curated chart-specific questions.

With the advent of representational learning over multi-modal data for document understanding \cite{appalaraju2021docformer}, \cite{gu2021unidoc}, \cite{zhong2019publaynet}, \cite{xu2020layoutlm}, \cite{powalski2021going}, \cite{li2021selfdoc}, the task of knowledge representation\cite{battaglia2018relational} and reasoning in latent space has improved significantly from previous heuristic-driven methods used to capture propositional logic. 

Recent works have paved the way for linking $FOC_{2}$ (First Order Logic with two variables and counting capability) with neural networks\cite{barcelo2020logical}. Tasks such as learning to reason over mathematical expressions \cite{9412619},\cite{mansouri2022advancing}  make a neat test bed for NSC(Neuro-Symbolic-Computing)\cite{nsc_survey}. 

There has been a rich history of research in building logic-based systems, for Theorem Proving, Conjecture Solving etc. Recent advances have seen the efficacy of using sophisticated  transformers and graph neural networks for the purpose of mathematical reasoning over very large datasets involving millions of intermediate logical steps\cite{kaliszyk2017holstep}. 

One recent work claims almost a $20\%$ jump in accuracy for synthetic chart QA, just by augmenting pretraining with mathematical reasoning \cite{liu2022matcha} although they lack a robust evaluation of the models' reasoning capability.

To further develop models capable of formal logic in the space of document understanding, we propose RealCQA as a robust multimodal testbed for logic and scientific chart-based QA. 

% \section{Contribution}
% \begin{itemize}
    
%     \item Benchmark 
%     \item Propose a  
% \end{itemize}

%% file: Sections/002_Background.tex
\section{Background}
We first discuss more commonly studied tasks in the literature that provide a foundation for ChartQA. These include visual QA, document understanding, and formal logic systems.
% % ----------------------------
\subsection{Visual QA}
VQA, or Visual QA, is a task where a computer system is given an image and a natural language question about the image and the system is expected to generate a natural language answer\cite{9813988}. VQA systems aim to mimic the ability of humans to understand and reason about visual information and language and to use this understanding to generate appropriate responses to questions. Specific variations of this task include image captioning and multi-modal retrieval. 
% VQA systems typically involve the integration of computer vision and natural language processing techniques and are used in a variety of applications such as image and video search, virtual assistants, and automated customer service.

%% ----------------------------
\subsubsection{Image Captioning} is the task  where a computer system is given an input image and is expected to generate a natural language description of the content of the image\cite{chen2019neural} . This description should capture the main objects, actions, and events depicted in the image, as well as the relationships between them. Image captioning systems typically use machine learning algorithms to learn how to generate descriptive captions from a large dataset of images and their corresponding human-generated captions. 
% Image captioning has many potential applications, including improving the accessibility of visual content for people with visual impairments, enabling the search and retrieval of images based on their content, and helping to automatically annotate and organize large collections of images.
% %%% ----------------------------
\subsubsection{Multimodal Retrieval}involves retrieving images and text that are related to each other based on their content\cite{Jawade_2023_WACV}. This task involves the integration of computer vision and natural language processing techniques and is used in a variety of applications such as image and text search, image annotation, and automated customer service. In image-text cross-modal retrieval, a computer system is given a query in the form of either an image or a text and is expected to retrieve images or texts that are related to the query. To perform image-text cross-modal retrieval effectively, the system must be able to understand and reason about the visual and linguistic content of both images and text and to identify the relationships between them. This typically involves the use of machine learning algorithms that are trained on large datasets of images and text and their corresponding relationships.

% %% ----------------------------

\subsection{Document Understanding} The field of document intelligence encompasses a broad range of tasks \cite{borchmann2021due}, \cite{tanaka2021visualmrc},  such as localization, recognition, layout understanding, entity recognition, and linking. In this section, we describe the downstream tasks of document-QA, Table-QA, and Infographic-QA, which build up to Chart-QA

% %%% ----------------------------
\subsubsection{VQA for Document Understanding} has been explored in works such as \cite{zhu2022towards_docvqa}  which involve document pages comprising of tables, text and QA-pairs. The documents are sampled from financial reports and contain lots of numbers, requiring discrete reasoning capability to answer  questions. Relational-VQA models use reasoning frameworks based on FOL to answer questions about visual scenes. Researchers have also explored other figure types such as Map-based QA \cite{chang2022mapqa}. CALM \cite{du2022calm} proposes extending \cite{mathew2021docvqa} with prior knowledge reasoning, and \cite{vsvcavnicka2022towards} proposes models for non-English document understanding through QA. Key requisites for Document-QA [DQA] include \textbf{(i)}\emph{Robust feature representation:} One of the main challenges in DQA is to effectively represent the visual and semantic content of documents. The development of robust feature representations that capture the relationships between objects, properties, and concepts in documents is a key area of research in the field. \textbf{(ii)}\emph{Large-scale datasets:} Another challenge in DQA is the lack of large-scale datasets that can be used to train and evaluate models. The development of large-scale datasets that include a wide variety of documents and questions is crucial for advancing the field. \textbf{(iii)}\emph{Integration of prior knowledge and context:} In order to accurately answer questions about documents, models must be able to effectively integrate prior knowledge and context into their reasoning process. This requires the development of algorithms that can reason about the relationships between objects and concepts in a document, and that can incorporate prior knowledge and context into the decision-making process. \textbf{(iv)}\emph{Relational reasoning:} DQA often requires reasoning about relationships between objects and concepts in a document. \textbf{(v)}\emph{Multi-modal fusion:} DQA requires the integration of information from multiple modalities, including visual and semantic content.
Recent works include \cite{mathew2021docvqa}, \cite{wu2022region}, \cite{tito2021icdar}, \cite{qi2022dureadervis}. 
% The development of algorithms that can perform relational reasoning is a key area of research in the field. 
% The development of algorithms that can effectively fuse information from multiple modalities is another key area of research.

% %%% ----------------------------
\subsubsection{Table QA} is a natural language processing (NLP) task that involves answering questions about the information presented in tables. This task requires models to understand the structure and content of the table, as well as the meaning of the natural language question, in order to generate a correct answer. The table contents are provided as text input. Recent literature in the TableQA task include \cite{herzig2020tapas}, \cite{mate_tab_qa}, which present models for generating SQL queries from natural language questions about tables. 

\subsection{Chart-VQA}
We discuss two common approaches for this specific sub-area of IQA where the input is a chart image and a corresponding query.

 \subsubsection{Semi-Structured Information Extraction (SIE)\cite{masry2022chartqa}} involves the following steps:\textbf{(i)}\emph{Chart Text Analysis:}  Extract the tick labels, legend, axis and chart tiles, and any other text in the image. \textbf{(ii)}\emph{Chart Structure Analysis:} Tick association for corresponding data value interpolation of $xy$ coordinate and the nearest tick label and legend mapping to individual data-series components labels. \textbf{(iii)}\emph{ Visual Element Detection [VED]:} Localize the chart component (line, box, point, bar) and association with x-tick and legend name. \textbf{(iv)}\emph{Data Extraction:} Interpolate the value represented by each data component by using the VED module and calculate the value from bounding ticks.

This reduces Chart-VQA to a Table-VQA task. However, this adds additional complexity as errors are now also introduced during the data-extraction task. 
%in addition to from the performance of visual and textual modality. 

% Overall, SIE is a useful technique for extracting data from plots and organizing it into a semistructured format that can be easily analyzed and processed. It involves a series of steps that involve extracting and organizing the relevant information, and requires the use of OCR and VED modules to extract the necessary data.

% Reference:

% Methani, N., et al. (2019). "Data Interpretation over Plots." arXiv preprint arXiv:1909.00997.
% %%% ----------------------------
\subsubsection{Classification-Regression\cite{levy2022classification}, \cite{methani2020plotqa}} approach has proven to be effective for chart comprehension, allowing machine learning models to accurately classify and predict the values and trends depicted in charts. In this school of thought, the input is directly treated as just pixels, usually relying on the implicit representation of chart components, plot area, visual elements, and underlying data.  These features are aggregated alongside text features of the question string  where the model learns their corresponding relations to predict either a classification answer(string) or a regression answer(numeric). Usually, models use visual features from a Mask-RCNN-based backbone, trained to detect chart text and structure. These are input alongside tokenized textual queries. Answer prediction involves predicting numeric or string type, where floats are regressed and tokens classified. 
% ---------------------

\begin{figure*}[ht!]
         \centering
         \includegraphics[scale=0.45]{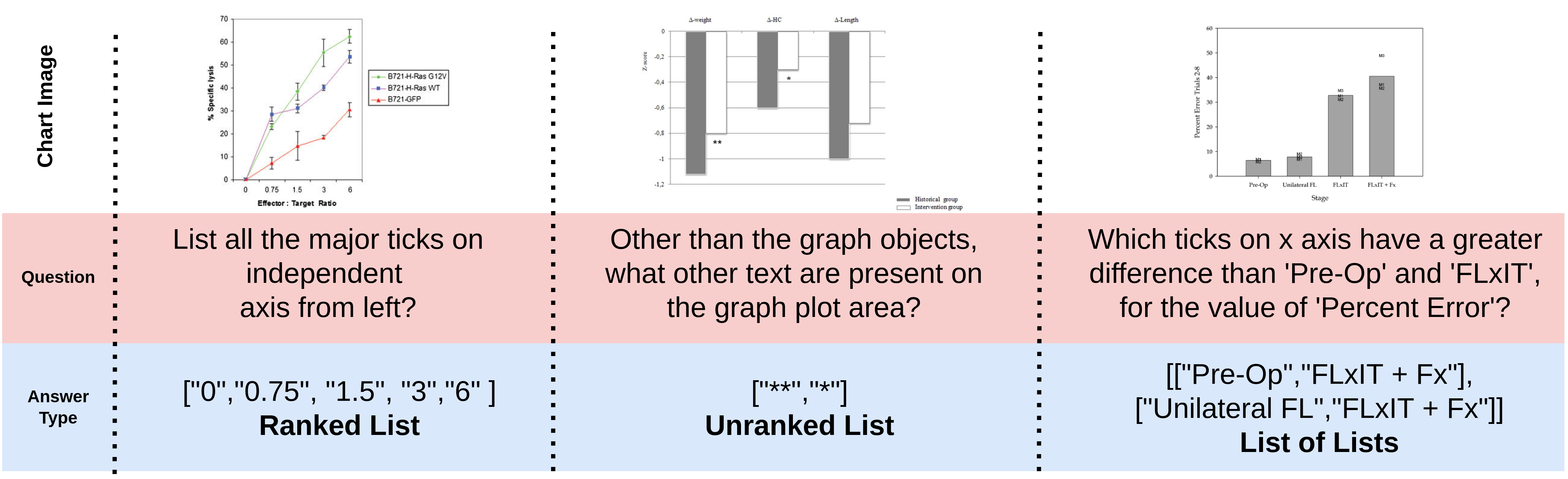}
         \caption{ \scriptsize List type answers have many uses-cases specifically in chart QA but has not been considered by existing CQA datasets. RealCQA introduces List type QA pairs with both (i) Ranked List (ii) Unranked List. Items in a list can consist of sets of up to 2 items.}
         \vspace{-1.2em}
         \label{fig:listdesc}
\end{figure*}

\subsection{Logic Order and Reasoning} 
We discuss formal logic, requirements for a testbed, and its applicability in the context of Chart-QA. 

\subsubsection{Zero-order logic [ZOL]} is the basic unit of meaning, the atomic formula, a proposition that makes an assertion,
\textit{e.g.} answering the root level taxonomy questions: \emph{`Is this chart of type A'}, \emph{`Is there a title in the chart'}, \emph{`Is the dependent axis logarithmic"} etc. Complex statements can be formed by combining atomic formulas using logical connectives such as `and', `or', and `not'. 

\subsubsection{First-order logic [FOL]} also known as predicate logic, is a type of formal logic that is used to study the relationships between objects and their properties. It allows the expression of propositions or statements that make assertions about properties and relations, and it provides a formal language for making logical inferences based on these assertions.
In first-order logic, we have quantifiers $\forall$ (for all) and  $\exists$ (there exist) to make statements about the entire domain of discourse by involving variables that range over the objects being discussed, \textit{e.g.} a closed set of all tick labels in a chart. We can talk about the properties of these objects or relationships between them by using 1-place predicates or multi-place predicates, respectively, \textit{e.g.} comparing data series values at different tick locations. These predicates can be viewed as sets where their elements are those of the domain that satisfy some property or n-tuples that satisfy some relation, \emph{`Is the sum of the value of $\langle$Y title$\rangle$ in $\langle$ i-th x tick$\rangle$ and $\langle$(i + 1)th x tick$\rangle$ greater than the maximum value of $\langle$ title$\rangle$ across all $\langle$plural form of X title$\rangle$ ?'}

 \subsubsection{N-th order logic} uses the same quantifiers to range over predicates. This essentially allows the quantification of sets. Using the quantifiers for elements of the sets is provisional as required. These involve the `List' type questions with structured output, \textit{e.g.} \emph{`Which pairs of major ticks on independent axis have a difference greater than $\langle$i-th x tick$\rangle$ and $\langle$j-th x tick$\rangle$, for the value of $\langle$Y title$\rangle$, arranged in the increasing order of difference?'}. Constraining the current scope for better evaluation, we limit this to $2^{nd}$ order logic, \textit{i.e.} we create questions with set outputs of at most 2 elements per item in the list as depicted in Fig. \ref{fig:listdesc}. 
% First-order logic is a formal system that uses predicates and quantifiers to represent relationships between objects and properties. In FOL, predicates can have one or more arguments, and quantifiers can be used to express the existence or non-existence of objects with certain properties. FOL is used in many branches of mathematics and computer science, including model theory, automated theorem proving, and knowledge representation.

% Zero-order logic, on the other hand, is a simpler form of logic that does not use quantifiers or predicates. Instead, it only uses propositional variables and logical connectives, such as AND, OR, NOT, and IMPLIES. ZOL is typically used to express simple statements and relationships between objects, without the need for a more complex representation.

% In summary, first-order logic is more expressive and powerful than zero-order logic, but also more complex. The choice of which logic to use depends on the requirements of the specific application and the desired level of expressiveness and complexity.
 \subsubsection{A Testbed for Formal Logic} must satisfy specific requirements to ensure the correctness of the system being tested. The first requirement is a formal specification, which should precisely define the system's syntax, semantics, and model-checking algorithms. The second requirement is a set of test cases that covers all possible scenarios and validates the system's behavior, including its ability to handle edge cases and exceptional conditions. The third requirement is a repeatable, reliable, and easily extensible test harness that can accommodate new test cases. Finally, a verification environment must be created to host the testbed and provide necessary resources for performing the tests.

 These ensure that formal logic systems are thoroughly tested and that the test results are accurate and reliable. The exact nature of the requirements and techniques used to meet them will vary depending on the system being tested and the context in which it is used. Overall, rigorous testing is crucial for establishing the correctness of logical systems and ensuring their applicability to real-world problems. Existing experimental setups for evaluating Chart-QA satisfy most of the above points, except the formal specification, which we provide for a subset of our questions. These are manually curated and verified. We will describe this in greater detail.
% %%% ----------------------------
\subsubsection{CQA for FOL} represents our concept of utilizing the template-based chart QA task as a testbed of predicate logic. The innate structure of data which populates a scientific chart aligns naturally with the previously stated formal specification requirements. Prior research has studied VQA with charts, however, a formal testbed has not been studied. 

 In FOL, sentences are written in a specific syntax and structure to allow for precise and unambiguous representation of meaning. To translate a normal sentence into FOL, we need to identify the objects, individuals, and relationships described in the sentence, and express them using predicates, variables, and logical connectives. 
 For example, the template \emph{Is the difference between the value of $\langle$ Y title $\rangle$ at $\langle$ ith x-tick $\rangle$ and $\langle$ jth x-tick $\rangle$ greater than the difference between any two $\langle$ plural form of X-title $\rangle$ ?}  can be converted to FOL as : 
 $$
 \forall i,j,p,q : (i \neq j \neq p \neq q) \rightarrow ( |Y_i-Y_j| > |X_p - X_q|) \;$$ where $\langle ... \rangle$ represents variables in the template and `Y' the space of values of the dependent-variable, and `X' for the independent variable in the chart, respectively.
 While curating reasoning type \cite{kafle2018dvqa} questions, we create a subset of binary questions specifically over FOL that are valid. 
 %(refer to supplementary for Full List).
 % We use this to benchmark publicly available Large Language Models.

In this study, our aim is to further advance the development of the chart and visual data parsing systems. Previous research has documented the limitations posed by the limited availability of annotated real-world data\cite{icpr2022}. While absolute accuracy on a specific ChartQA dataset may not guarantee broad generalizability, this study is a step toward establishing a comprehensive understanding of this complex and evolving field. We believe that leveraging the manually curated templates and structured output generated from the semantic structure of charts presents an opportunity to effectively evaluate the multi-modal predicate logic parsing capabilities of modern neural networks, such as large-scale pre-trained language and layout models.

%% file: Sections/003_Methodology.tex
\section{Dataset}

\input{Sections/Figures/03_icpravlabl.tex}

In this section, we describe the dataset used in our study. The dataset, called RealCQA, was created by utilizing real-world chart images and annotations used in publicly conducted chart understanding challenges \cite{icpr2022}. Fig.\ref{fig:existDatase} shows the current existing datasets in the CQA domain. The challenge tasks around chart understanding are shown in Fig. \ref{fig:challenge_data}, along with the annotated data used from the publicly released train-test splits.
\subsection{RealCQA}
To generate question templates for RealCQA, we compiled templates from previous works \cite{chaudhry2020leaf}, \cite{singh2020stl}, and \cite{methani2020plotqa}. These templates were adapted to our data and augmented with new chart-type questions, list questions, and binary FOL reasoning questions, forming a total of 240 templates. 

The distribution of taxonomy and answer types for RealCQA is shown in Fig. \ref{fig:piedist}. We have tried to keep the templates for different answer types with equal proportions. However, when these are used to create the actual QA pairs, the data gets skewed depending on underlying availability. Our dataset consists of a majority of `Reasoning' type questions, as seen in previous works \cite{kafle2018dvqa}. However, we also focus on creating binary reasoning questions that satisfy FOL. These form a major chunk of the dataset since templates with variables for i-th/j-th tick/data series are combinatorial in nature, and we create them exhaustively over the closed set of objects present in the chart.

We use the `Structure, Retrieval, Reasoning' taxonomy proposed in previous works \cite{kafle2018dvqa} to categorize our questions. However, we further demarcate them as Types 1, 2, 3, 4, depending on their characteristics. Type-1 refers to any questions that can be formed at the (root) level of the whole chart image, mostly ZOL. Type-2 further refers to ZOL questions for specific chart components, requiring the model to identify them. Type-3 and Type-4 are data retrieval/reasoning. Each has a further specific sub-class depending on the exact component, chart type, etc., as shown in Fig. \ref{fig:piedist}. 

The statistics of the dataset are shown in Fig. \ref{fig:dataStats} using the previous nomenclature of `Structural', `Retrieval', and `Reasoning'. For the List Type, we only curate reasoning questions for $k^{th}$ order FOL testing. String/Unranked refers to a small subset of string-type retrieval or reasoning answers where multiple equivalent conditions exist. While reading the question string, a human would expect a single answer, but multiple data series have the same maximum/minimum, resulting in multiple correct single-string instance answers. These are generally outliers. 

Overall, the RealCQA dataset offers a diverse range of questions that require various levels of chart understanding and reasoning abilities. The dataset is publicly available and can be used for further research and evaluation in the chart understanding domain.

\input{Sections/Figures/01_tax_dist.tex}

\input{Sections/Figures/02_train_test_stat.tex}

\input{Sections/Figures/04_sampl.tex}

\input{Sections/Tables/001.tex}

\subsection{Sampling Strategies for Dataset Evaluation}

For the purpose of general chart visual question answering, generating balanced and representative datasets is of paramount importance for training and evaluating models. However, when it comes to logic testbeds, the over-representation of specific templates or question types is not necessarily a disadvantage, as it allows for a more nuanced assessment of a model's logical reasoning capabilities. Nonetheless, it is still relevant to explore the impact of dataset sampling on evaluation results.

To this end, we devised five different sampling strategies and evaluated their effect on our dataset. The first strategy, exhaustive sampling, consists of including all available question-answer pairs. The remaining strategies aim to modify the distribution of questions per chart based on different criteria. Specifically, the second strategy, increasing lower bound, focuses on charts with a minimum number of questions greater than or equal to a threshold $K$. This strategy aims to address under-represented question types, such as root and structural questions. Conversely, the third strategy, decreasing upper bound, selects charts with a maximum number of questions less than or equal to a threshold $L$. This strategy is intended to address over-represented question types, typically combinatorial binary reasoning questions. The fourth strategy combines the effects of the second and third strategies, aiming to remove both under and over-represented charts. Finally, the fifth strategy, flat cap, selects a fixed number of questions per chart per template, thereby creating a more uniform dataset.

Fig. \ref{fig:sampling} illustrates how the different sampling strategies affect the number of questions per chart per template, while Table \ref{Tab:sample} provides the actual number of QA-pairs per sampling. To be specific, we calculate the lower and upper 10\% for the second and third strategies, respectively. For the flat cap strategy, we randomly select 150 QA-pairs for each template per chart.

By analyzing the impact of the different sampling strategies, we can gain a better understanding of how removing specific sections from the test-set affects evaluation results. These findings can be useful for modulating the training-set as required, and ultimately for developing more robust and accurate visual question-answering models.

% \subsection{Evaluation}

\subsection{Evaluation Metrics}

In this study, we propose an evaluation metric based on the accuracy of answers. The proposed task involves four types of answers, each with its specific calculation method.

First, for numerical answers, we measure the accuracy of regression errors using L2 or L1 differences or the ER-error rate. In PlotQA-D [1], we consider a regression answer correct if it falls within $\pm5\%$ tolerance from the ground truth value. Second, for single string answers, we use string-matching edit distance and count perfect matches as correct. Third, for unordered lists of strings, we use string-matching edit distance. For each of the $K$ queries and $M$ matches, we calculate $K \times M$ scores, and the mutually exclusive best match is aggregated per string instance normalized by $K$. Fourth, for ranked order lists of strings, we use the nDCG@K ranking metric, where K is the size of the ground-truth list. nDCG is a normalized version of the DCG (Discounted Cumulative Gain) metric, which is widely used to evaluate the ranking quality of information retrieval systems, such as search engines and recommendation systems. This metric assigns a relevance score to each item in a ranked list based on the user's preferences, and then discounts these scores using a logarithmic function, with items appearing lower in the ranking receiving lower scores. Lastly, for nested lists, where each item is a set, we evaluate the results invariant of set order, but list order matters in ranked lists.
\section{Experiments}

We benchmark multiple existing generic visual QA and chart-specific visual QA methods on RealCQA. Fig \ref{fig:piline} shows the generic existing architecture used for CQA task. The model either learns visual and data features separately or in the same shared space and then uses some fusion model to generate the final answer. These are primarily trained on synthetic charts. Here, we evaluate multiple baseline models that have been proposed recently including ChartQA and CRCT. We present both the synthetic pre-training evaluation on RealCQA and RealCQA finetuned evaluation. Below we briefly discuss the model architecture for the baseline methods in more detail:

\begin{figure*}[ht!]
         \centering
         \includegraphics[scale=0.3]{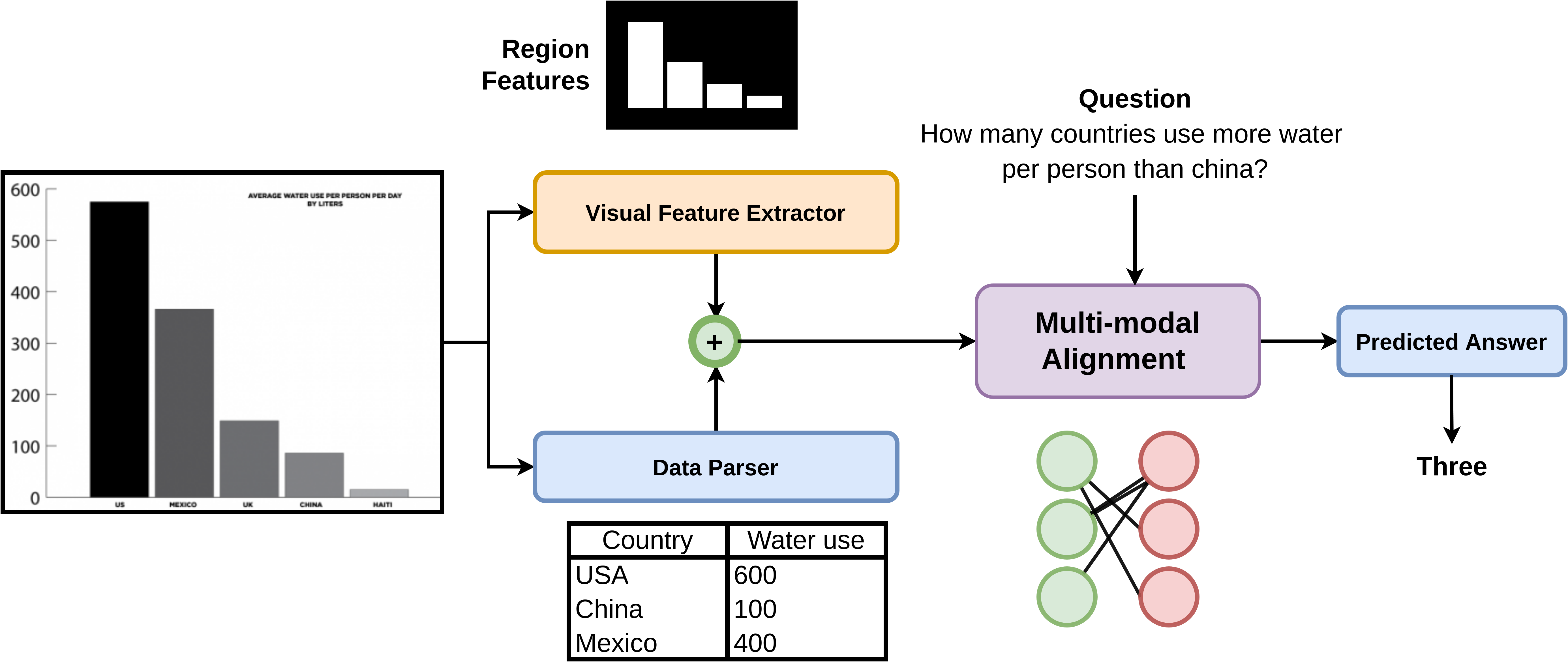}
         \caption{A generalized framework to represent existing methods for chart QA.}
         \vspace{-1.2em}
         \label{fig:piline}
\end{figure*}

\subsubsection{VLT5\cite{cho2021unifying}}
VLT5 is a state-of-the-art unified framework that leverages a multimodal text conditioning language objective to perform different tasks within a unified architecture. In this framework, the model learns to generate labels in the text space based on the visual and textual inputs. In our study, we use VLT5 to perform the task of table-based question answering. Specifically, VLT5 takes as input pre-trained region-based visual features obtained from Faster-RCNN, which was pre-trained on PlotQA \cite{methani2020plotqa}. These visual features, along with textual tokens, are projected and fed through a unified bi-directional multi-modal encoder. Additionally, a language decoder is trained in an auto-regressive setting to perform text generation. In the textual context, we provide a pre-extracted gold standard table of the chart as concatenated input along with the query question.

We present the performance of VLT5 on the RealCQA test-set, which comprises approximately 683 charts with gold-data table annotation. The evaluation is conducted on the charts in this test-set, with a score of zero assigned to the remaining charts. The results of this evaluation are presented in Table 2 and Table 3, where Row 1 shows the performance of VLT5 segregated by Answer-Type and Question-Type, respectively.
% \subsubsection{ChartQA \cite{masry2022chartqa}} 
% proposed a large-scale benchmark dataset with 9.6K human-written questions as well as 23.1K questions generated from human-written chart summaries. The paper
% benchmarked two transformer based multi-modal architectures namely VisionTapas and VLT5 on ChartQA dataset using data table and visual features as context. We fine-tune the VLT5 multi-modal encoder with ChartQA visual features pre-trained on PlotQA on RealCQA.  Both Row 1 and Row 2 in Table 2 and Table 3 utilize ChartQA pre-trained Mask RCNN visual features with VLT5 multi-modal attention. Results are segregated on basis of Answer-Type and Question-Type in Table 2 and Table 3 respectively. As previously, since this evaluation requires a data-table, we evaluated it for 683 charts with zero score assigned to remaining charts QAs.

\subsubsection{ChartQA \cite{masry2022chartqa}} 
The authors of ChartQA introduced a large-scale benchmark dataset comprising of 9.6K human-written questions and 23.1K questions generated from human-written chart summaries. To evaluate the effectiveness of the dataset, two transformer-based multimodal architectures, namely VisionTapas and VLT5, were benchmarked on ChartQA using data tables and visual features as context. 

In this study, we fine-tuned the VLT5 multi-modal encoder with ChartQA visual features that were pre-trained on PlotQA. We utilized ChartQA pre-trained Mask RCNN visual features with VLT5 multi-modal attention in both Row 1 and Row 2 of Table 2 and Table 3, respectively. Results were segregated based on Answer-Type and Question-Type, as presented in Table 2 and Table 3. Since the evaluation requires a data-table, we evaluated the model on 683 charts and assigned zero scores to the remaining QAs. 

% \subsubsection{CRCT \cite{levy2022classification}} 
% % proposed a novel ChartVQA approach, the Classification Regression Chart Transformer (CRCT), to address the limitations of existing methods. The authors argue that the saturation of previous methods is due to biases, over-simplicity, and classification-oriented Q\&A in common datasets and benchmarks. To tackle this challenge, the CRCT model leverages a dual branch transformer with a chart element detector to extract both textual and visual information from charts. The model also features joint processing of all textual elements in the chart to capture inter and intra-relations between elements. They 
% introduced a co-transformer that fundamentally is a multi-modal pre-trained BERT that allows to fuse both visual and textual information into a pooled tuple of two single feature vectors. Additionally, the model introduces a new chart element representation learning mechanism that fuses multiple inputs from different domains. The proposed hybrid prediction head unifies classification and regression into a single model, optimizing the end-to-end approach using multi-task learning. For visual context they fine-tuned a Mask-RCNN on PlotQA and textual context they use text detections and recognition output from a standard OCR (such as tesseract). We evaluate both a CRCT pretrained fully pre-trained on PlotQA dataset and CRCT fine-tuned on RealCQA for stage 2 with pre-trained FasterRCNN and report the performance (row 3, 4) in Table 2 and Table 3.
\subsubsection{CRCT \cite{levy2022classification}} 
The paper proposes a novel ChartVQA approach called Classification Regression Chart Transformer (CRCT) that aims to address the limitations of existing methods in the field. The authors argue that the saturation of previous methods is due to biases, oversimplification, and classification-oriented Q\&A in common datasets and benchmarks. To overcome these challenges, the CRCT model leverages a dual-branch transformer with a chart element detector that extracts both textual and visual information from charts. The model also features joint processing of all textual elements in the chart to capture inter and intra-relations between elements. 

% The authors introduced a co-transformer, which is a multi-modal pre-trained BERT that fuses both visual and textual information into a pooled tuple of two single feature vectors. Additionally, the model proposes a new chart element representation learning mechanism that fuses multiple inputs from different domains. 
The proposed hybrid prediction head unifies classification and regression into a single model, optimizing the end-to-end approach using multi-task learning. For visual context, they fine-tuned a Mask-RCNN on PlotQA, while for textual context, they used text detections and recognition output from a standard OCR such as tesseract. We evaluated both a CRCT model fully pre-trained on the PlotQA dataset and a CRCT model fine-tuned on RealCQA for stage 2 with pre-trained FasterRCNN. We report the performance of both models in Row 3 and Row 4 of Table 2 and Table 3, respectively.

\subfile{./Sections/Tables/answertypebaselines}
\subfile{./Sections/Tables/questiontypebaseline}

\subsection{Results}

We present the quantitative results of our experiments, which are summarized in Table 2 and Table 3. We find that the VLT5 model \cite{cho2021unifying}, does not perform well on the RealCQA dataset when using pre-trained Mask RCNN visual features and RealCQA's gold-data table as input. The performance of 49.16\% on binary-type answers is as bad as random assignment. However, by fine-tuning the VLT5 model's multi-modal alignment module on RealCQA with modified tokenization to handle list-type answers, we observe significant improvements in performance on all answer types (Table 1) and question types (Table 2). Specifically, the performance on string type answers improves from 0.008\% to 30.68\%, and the overall accuracy of QA pairs improves from 23.99\% to 31.06\%. Table 3 compares the performance of the CRCT model \cite{levy2022classification} fully pre-trained on PlotQA and VG Pretrained VLT5 on Root, Structure, and Retrieval type questions. We find that the fully pre-trained CRCT outperforms VG Pretrained VLT5 on these question types. However, fine-tuning the CRCT model on RealCQA leads to significant improvements in performance on Numeric, Ranked List, and Unranked List type answers, as shown in Table 2. Notably, fine-tuned CRCT achieves the best performance of 31.58\% on numeric type answers. Our results highlight the importance of RealCQA as a standard test bed for evaluating chart visual QA methods, as even models that perform well on synthetic datasets such as PlotQA and FigureQA struggle to generalize to real-world chart distributions.

\input{Sections/Tables/002.tex}

 We present an ablation study that examines the impact of different sampling strategies on the performance of our model. The results of this study are summarized in Table \ref{tab:tablelabeling}. The study reveals that the 4th sampling strategy, which combines both the upper and lower bounds, consistently achieves the highest overall, string, and binary type accuracy across various experimental settings. On the other hand, the 5th strategy, which produces the most uniform test set, yields top accuracies for numeric and list-type answers. However, this strategy has the smallest size in terms of overall, unranked, and binary questions, and it attains the highest accuracy for unranked list type questions, which are representative of Kth Order Logic questions. It is worth noting that the 5th strategy may remove most of the challenging QA pairs, making it less desirable for our objective.

Overall, this study highlights the importance of carefully selecting the sampling strategy to obtain a test set that is representative of the distribution of real-world chart visual QA. The 4th strategy appears to be a promising choice, as it achieves high accuracy across different answer types while maintaining a sufficient number of challenging QA pairs.

%% file: Sections/Figures/03_icpravlabl.tex
\begin{figure}[ht!]
     \centering
     \begin{subfigure}[b]{0.45\textwidth}
         \centering
         \includegraphics[width=\textwidth]{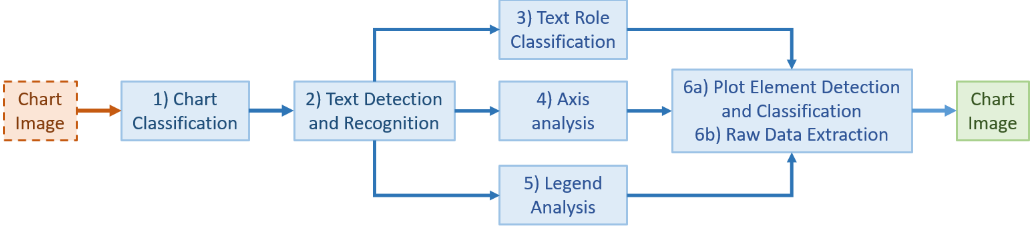}
         \caption{Challenge Tasks}
         \label{fig:tax_dist}
     \end{subfigure}
     \hfill
     \begin{subfigure}[b]{0.4\textwidth}
         \centering
         \includegraphics[width=\textwidth]{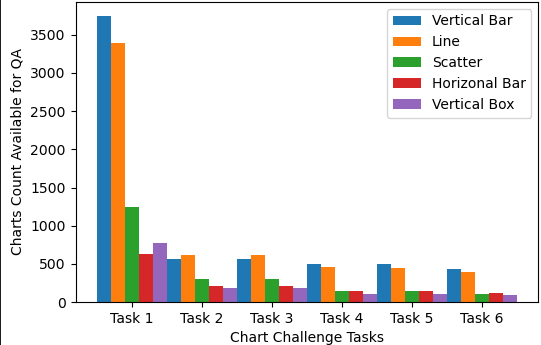}
         \caption{Data Distribution}
         \label{fig:tid_dist}
     \end{subfigure}
     \hfill
        \caption{ \scriptsize Train and Test Structure, Retrieval, Reasoning by answer type. For List Type, we only curate reasoning questions for kth order FOL testing. String/Unranked refers to a small subset of string-type retrieval or reasoning answers where multiple equivalent conditions exist: While reading the question string, a human would expect a single answer but multiple data series have the same maximum/minimum etc. resulting in multiple correct single-string instance answers. }
        \label{fig:challenge_data}
\end{figure}

%% file: Sections/Figures/01_tax_dist.tex
\begin{figure}[ht!]
         \centering
         \includegraphics[width= \textwidth]{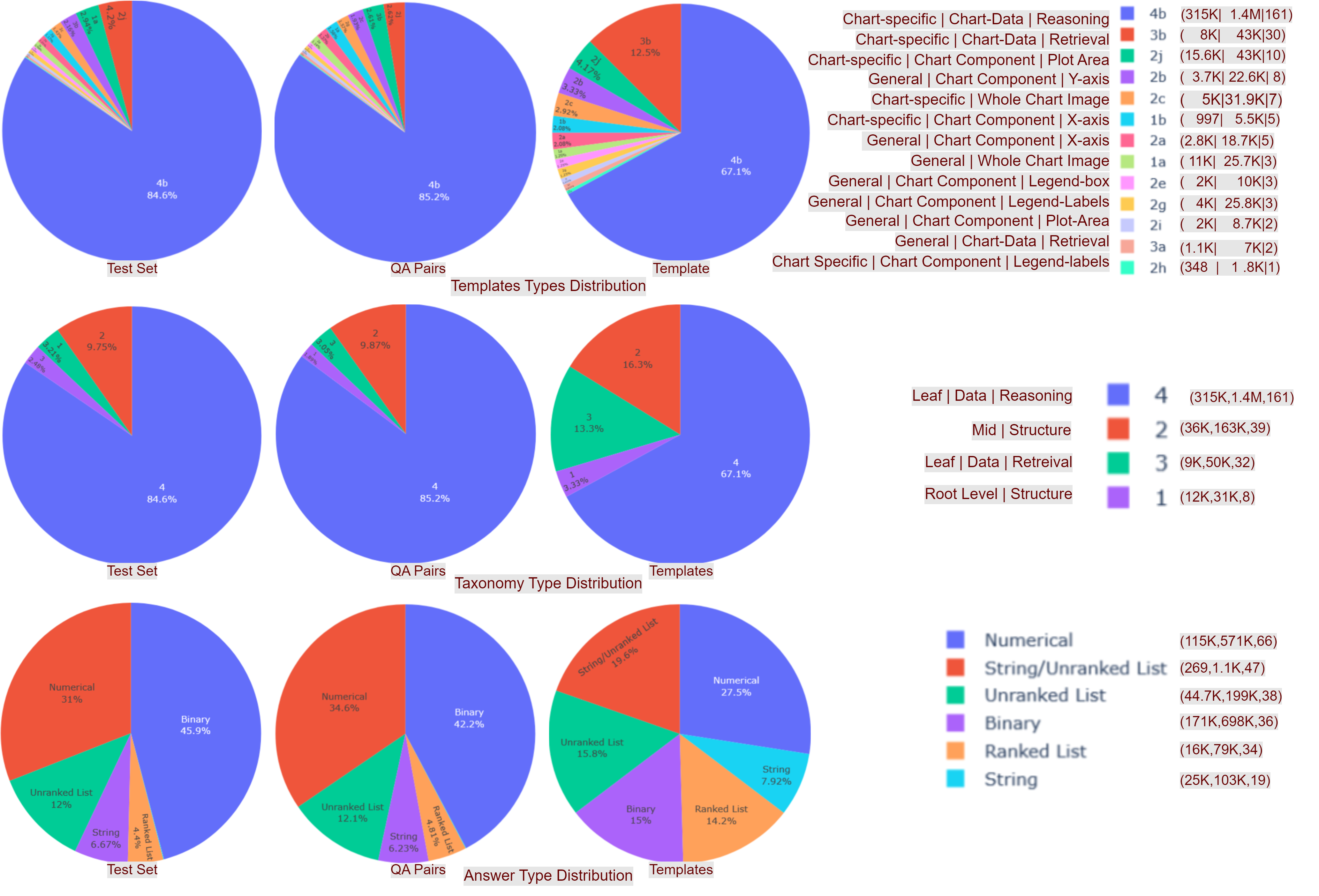}
         \caption{\scriptsize Taxonomy and Type distribution: First pie is test set QA pairs, second training set QA pairs and third training set templates. Legend is in decreasing order. Best viewed digitally in color.}
         \label{fig:piedist}
\end{figure}

%% file: Sections/Figures/02_train_test_stat.tex
\begin{figure}[ht!]
     \centering
     \begin{subfigure}[b]{0.465\textwidth}
         \centering
         \includegraphics[width=\textwidth]{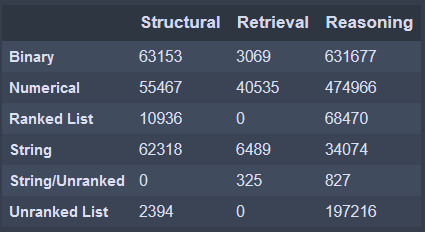}
         \caption{Train Distribution}
         
     \end{subfigure}
     \hfill
     \begin{subfigure}[b]{0.475\textwidth}
         \centering
         \includegraphics[width=\textwidth]{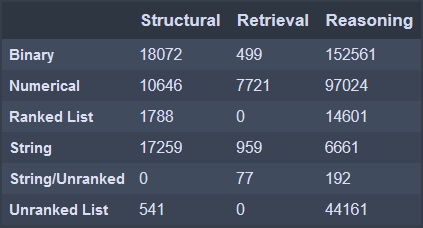}
         \caption{Test Distribution}
     \end{subfigure}
     \hfill
        \caption{ \scriptsize Train and Test Structure, Retrieval, Reasoning by answer type.  }
        \label{fig:dataStats}
\end{figure}

%% file: Sections/Figures/04_sampl.tex
\begin{figure}[ht!]
         \centering
         \includegraphics[width= \textwidth]{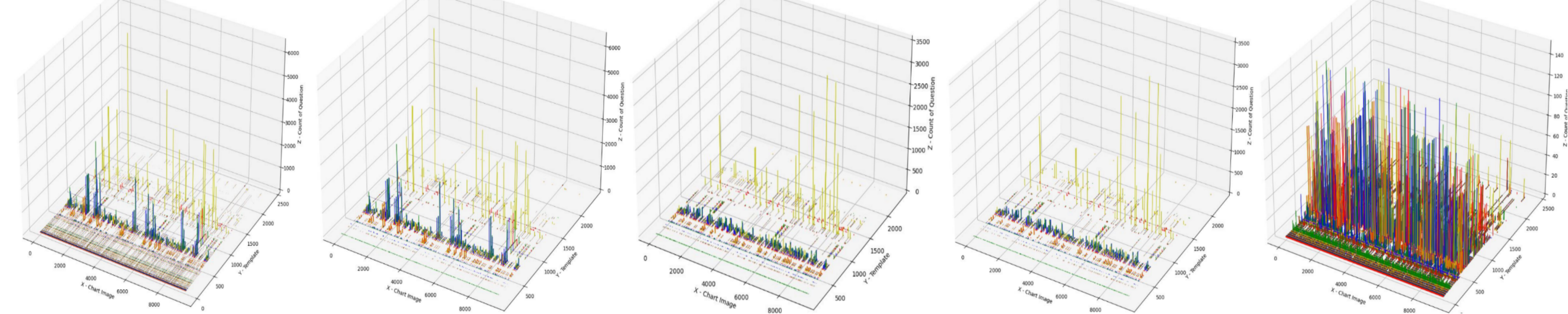}
         \caption{\scriptsize Trend across different sampling strategies 1-5. X-axis represents each of the 9357 test-images, Y-axis the 240 templates each plotted with different colored bars at every 10th index, and Z axis shows the count of QA pairs.}
         \label{fig:sampling}
\end{figure}

%% file: Sections/Tables/001.tex
\begin{table}[ht!]
\centering
\scriptsize
\caption{\scriptsize Total QA pairs per sampling strategy, bold shows minimum.}
\begin{tabular}{|c|ccccc}
\hline
Answer Type &
  \multicolumn{1}{c|}{Sample 1} &
  \multicolumn{1}{c|}{Sample 2} &
  \multicolumn{1}{c|}{Sample 3} &
  \multicolumn{1}{c|}{Sample 4} &
  \multicolumn{1}{c|}{Sample 5} \\ \hline
Total    & 367139 & 322404 & 276091 & 231356         & \textbf{203735} \\ \cline{1-1}
String   & 19525  & 3489   & 18548  & \textbf{2512}  & 19046           \\ \cline{1-1}
Numeric  & 115391 & 107153 & 93096  & \textbf{84858} & 78680           \\ \cline{1-1}
Ranked   & 16389  & 13903  & 13357  & \textbf{10871} & 14228           \\ \cline{1-1}
Unranked & 44702  & 43310  & 27019  & 25627          & \textbf{25041}  \\ \cline{1-1}
Binary   & 171132 & 154549 & 124071 & 107488         & \textbf{66740}  \\ \cline{1-1}
\end{tabular}
\label{Tab:sample}
\end{table}

%% file: Sections/Tables/answertypebaselines.tex
\begin{table}[h]
\label{TableQuestionType}
\centering
\caption{Performance of existing Visual Question Answering Methods on RealCQA based on Answer Type}
\vspace{1.0em}
\begin{tabular}{
>{\columncolor[HTML]{FFFFFF}}c |
>{\columncolor[HTML]{FFFFFF}}c |
>{\columncolor[HTML]{FFFFFF}}c |
>{\columncolor[HTML]{FFFFFF}}c |
>{\columncolor[HTML]{FFFFFF}}c |
>{\columncolor[HTML]{FFFFFF}}c |
>{\columncolor[HTML]{FFFFFF}}c }
\hline
\multicolumn{1}{l|}{\cellcolor[HTML]{FFFFFF}\textbf{}}                                       & {\color[HTML]{1D1C1D} \textbf{\begin{tabular}[c]{@{}c@{}}Total\\ Accuracy\end{tabular}}} & {\color[HTML]{1D1C1D} \textbf{String}} & {\color[HTML]{1D1C1D} \textbf{Numeric}} & {\color[HTML]{1D1C1D} \textbf{Rank}} & {\color[HTML]{1D1C1D} \textbf{Unranked}} & {\color[HTML]{1D1C1D} \textbf{Binary}} \\ \hline
\begin{tabular}[c]{@{}c@{}}VLT5\\ (VG Pretrained)\end{tabular}                                  & {\color[HTML]{1D1C1D} 0.2399}                                                   & 0.0008                                 & 0.0325                                  & 0.0106                               & 0.0002                                   & 0.4916                        \\
{\color[HTML]{1D1C1D} \begin{tabular}[c]{@{}c@{}}VLT5\\ (RealCQA Finetuned)\end{tabular}} & \textbf{0.3106}                                                                                   & \textbf{0.3068}                                 & 0.1487                                  & 0.0246                               & 0.0048                                   & \textbf{0.5275}                                 \\
\begin{tabular}[c]{@{}c@{}}CRCT\\ (PlotQA Pretrained)\end{tabular}                                  & {\color[HTML]{1D1C1D} 0.1787}                                                            & {\color[HTML]{1D1C1D} 0.0350} & {\color[HTML]{1D1C1D} 0.0412}           & {\color[HTML]{1D1C1D} 0.0015}        & {\color[HTML]{1D1C1D} 0.0016}            & {\color[HTML]{1D1C1D} 0.3515}      \\
\begin{tabular}[c]{@{}c@{}}CRCT\\ (RealCQA Finetuned)\end{tabular}                           & {\color[HTML]{1D1C1D} 0.1880}                                                            & 0.0323                                 & \textbf{0.3158}                         & \textbf{0.0286}                      & \textbf{0.0124}                          & 0.1807                                 \\ \hline
\end{tabular}
\vspace{1em}

\end{table}

%% file: Sections/Tables/questiontypebaseline.tex
\begin{table}[]
\label{TableAnswerType}
\centering
\caption{Performance of existing Visual Question Answering Methods on Real-CQA based on Question Complexity}
\vspace{1.0em}
\begin{tabular}{
>{\columncolor[HTML]{FFFFFF}}c |
>{\columncolor[HTML]{FFFFFF}}c |
>{\columncolor[HTML]{FFFFFF}}c |
>{\columncolor[HTML]{FFFFFF}}c |
>{\columncolor[HTML]{FFFFFF}}c }
\hline
\multicolumn{1}{l|}{\cellcolor[HTML]{FFFFFF}\textbf{}}                                        & {\color[HTML]{1D1C1D} \textbf{\begin{tabular}[c]{@{}c@{}} \ Root \ \end{tabular}}} & \textbf{Structure}                            & \textbf{Retrieval}                            & \textbf{Reasoning}                           \\ \hline
\begin{tabular}[c]{@{}c@{}}VLT5 \\ (VG Pretrained)\end{tabular}                                  & {\color[HTML]{1D1C1D}  0.0764}                                                              & 0.1416                                        & 0.0765                                        & 0.2620                                       \\
{\color[HTML]{1D1C1D} \begin{tabular}[c]{@{}c@{}}VLT5 \\ (RealCQA Finetuned)\end{tabular}} & \textbf{0.1800}                                                                                   & \textbf{0.4352} & \textbf{0.5877} & \textbf{0.2937} \\
\begin{tabular}[c]{@{}c@{}}CRCT\\ (PlotQA Pretrained)\end{tabular}                                   & {\color[HTML]{1D1C1D} 0.0773}                                                            & 0.2338                                        & 0.1168                                        & 0.1778                                       \\
\begin{tabular}[c]{@{}c@{}}CRCT\\ (RealCQA Finetuned)\end{tabular}                            & {\color[HTML]{1D1C1D} 0.0115}                                                            & 0.1497                                        & 0.3131                                        & 0.1959                                       \\ \hline
\end{tabular}
\end{table}

%% file: Sections/Tables/002.tex
% \usepackage{multirow}
\begin{table}[ht!]
\scriptsize
\centering
\caption{An ablation on the performance of different models based on different sampling strategies. Here, LB - Lower Bound, UB - Upper Bound, Full - No Sampling.}
\begin{tabular}{l|c|c|c|c|c|c|c}
\hline
Data   & Model                                                                                  & \begin{tabular}[c]{@{}c@{}}Total \\ Accuracy\end{tabular} & String & Numeric & Rank   & Unrank & Binary \\ \hline
% \rowcolor{blue!10}

Full   & \multirow{5}{*}{\begin{tabular}[c]{@{}c@{}}VLT5\\ Pre-Trained\end{tabular}}            & 0.2399                                                    & 0.0008 & 0.0325  & 0.0106 & 0.0002 & 0.4916 \\ \cline{1-1} \cline{3-8} 
LB     &                                                                                        &                                                           & 0.0032 & 0.0252  & 0.0092 & 0.0001 & 0.5182 \\ \cline{1-1} \cline{3-8} 
UB     &                                                                                        &                                                           & 0.0008 & 0.0345  & 0.0090 & 0.0002 & 0.4947 \\ \cline{1-1} \cline{3-8} 
LB+UB  &                                                                                        &                                                           & 0.0044 & 0.0255  & 0.0069 & 0.0001 & 0.5334 \\ \cline{1-1} \cline{3-8} 
150max &                                                                                        &                                                           & 0.0008 & 0.0403  & 0.0122 & 0.0003 & 0.4349 \\ \hline
% \rowcolor{blue!10}
\\ \hline
Full   & \multirow{5}{*}{\begin{tabular}[c]{@{}c@{}}VLT5\\ (RealCQA\\ Fine-Tuned)\end{tabular}} & 0.3106                                                    & 0.3068 & 0.1487  & 0.0246 & 0.0048 & 0.5275 \\ \cline{1-1} \cline{3-8} 
LB     &                                                                                        &                                                           & 0.4004 & 0.1305  & 0.0201 & 0.0041 & 0.5487 \\ \cline{1-1} \cline{3-8} 
UB     &                                                                                        &                                                           & 0.3165 & 0.1585  & 0.0246 & 0.0070 & 0.5258 \\ \cline{1-1} \cline{3-8} 
LB+UB  &                                                                                        &                                                           & 0.5084 & 0.1365  & 0.0187 & 0.0061 & 0.5559 \\ \cline{1-1} \cline{3-8} 
150max &                                                                                        &                                                           & 0.3146 & 0.1827  & 0.0283 & 0.0085 & 0.4947 \\ \hline
% \rowcolor{red!10}
\\
\\\hline
Full   & \multirow{5}{*}{\begin{tabular}[c]{@{}c@{}}CRCT\\ Pretrained\end{tabular}}             & 0.1787                                                    & 0.0350 & 0.0412  & 0.0015 & 0.0000 & 0.3515 \\ \cline{1-1} \cline{3-8} 
LB     &                                                                                        &                                                           & 0.0224 & 0.0359  & 0.0013 & 0.0000 & 0.3473 \\ \cline{1-1} \cline{3-8} 
UB     &                                                                                        &                                                           & 0.0367 & 0.0428  & 0.0012 & 0.0000 & 0.3742 \\ \cline{1-1} \cline{3-8} 
LB+UB  &                                                                                        &                                                           & 0.0299 & 0.0363  & 0.0009 & 0.0000 & 0.3716 \\ \cline{1-1} \cline{3-8} 
150max &                                                                                        &                                                           & 0.0359 & 0.0499  & 0.0017 & 0.0000 & 0.3601 \\ \hline
\\% \rowcolor{red!10}
\hline
Full   & \multirow{5}{*}{\begin{tabular}[c]{@{}c@{}}CRCT\\ (RealCQAFine-Tuned)\end{tabular}}    & 0.1880                                                    & 0.0323 & 0.3158  & 0.0286 & 0.0124 & 0.1807 \\ \cline{1-1} \cline{3-8} 
LB     &                                                                                        &                                                           & 0.0759 & 0.3196  & 0.0263 & 0.0102 & 0.1923 \\ \cline{1-1} \cline{3-8} 
UB     &                                                                                        &                                                           & 0.0319 & 0.3316  & 0.0327 & 0.0190 & 0.1977 \\ \cline{1-1} \cline{3-8} 
LB+UB  &                                                                                        &                                                           & 0.0903 & 0.3379  & 0.0309 & 0.0158 & 0.2170 \\ \cline{1-1} \cline{3-8} 
150max &                                                                                        &                                                           & 0.0322 & 0.3371  & 0.0329 & 0.0220 & 0.1316 \\ \hline
\end{tabular}
\label{tab:tablelabeling}
\end{table}

%% file: Sections/004_Results.tex
\section{Conclusion}
% We have described the task of visual question answering on real-world charts. We have demonstrated the equivalence between solving a template-based Chart-QA problem and that of formal first-order logic in the domain of multimodal learning. We have described the construction of formal FOL queries from chart questions, further expanding research into multimodal fusion of image, text, and reasoning. 

In addition to our contribution of curating a novel FOL-Testbed and a dataset for the evaluation of CQA for real charts, we have also thoroughly evaluated several state-of-the-art visual question answering models on the RealCQA dataset. Our experiments reveal that while some models perform well on synthetic datasets like PlotQA and FigureQA, their performance significantly drops when tested on RealCQA, demonstrating the need for a more realistic and challenging benchmark like RealCQA. We have shown that our proposed method, CRCT, significantly outperforms previous models on several question types, especially on numeric type questions. Our ablation study further highlights the importance of sampling strategies in constructing a diverse and representative test set. 

Overall, our study emphasizes the importance of multimodal learning and reasoning in visual question answering and provides insights into the limitations and opportunities of current state-of-the-art models. Future work can build on our findings by exploring more sophisticated models that integrate text, image, and reasoning more effectively, as well as developing new evaluation metrics that capture the full complexity of real-world chart questions. Additionally, expanding the dataset to cover a wider range of chart types and complexities can further improve the generalization capabilities of visual question answering models and lead to more impactful applications in areas such as data analysis and decision-making.